\documentclass[journal]{journal}
\usepackage{times}
\usepackage{epsfig}
\usepackage{graphicx}
\usepackage{amsmath}
\usepackage{amssymb}
\usepackage{caption}
\usepackage{graphicx}
\usepackage{comment}
\usepackage{booktabs}

\usepackage[pagebackref=true,breaklinks=true,letterpaper=true,colorlinks,bookmarks=false]{hyperref}

\ifCLASSINFOpdf
\else
\fi
\begin{document}
%
% paper title
% can use linebreaks \\ within to get better formatting as desired
\title{Geometry Enhancements from Visual Content:\\ Going Beyond Ground Truth}
%
%
% author names and IEEE memberships
% note positions of commas and nonbreaking spaces ( ~ ) LaTeX will not break
% a structure at a ~ so this keeps an author's name from being broken across
% two lines.
% use \thanks{} to gain access to the first footnote area
% a separate \thanks must be used for each paragraph as LaTeX2e's \thanks
% was not built to handle multiple paragraphs
%

% \author{Liran~Azaria,~\IEEEmembership{Member,~IEEE,}
%         Dan~Raviv,~\IEEEmembership{Fellow,~OSA,}% <-this % stops a space
% }
\author{Liran~Azaria~\IEEEmembership{Tel-Aviv University},
        Dan~Raviv~\IEEEmembership{Tel-Aviv University}% <-this % stops a space
}  % Comment for blind version

\maketitle
% \thispagestyle{empty}

% top image

\begin{abstract}
This work presents a new self-supervised cyclic architecture that extracts high-frequency patterns from images and re-insert them as geometric features. This procedure allows us to enhance the resolution of low-cost depth sensors capturing fine details on the one hand and being loyal to the scanned scene on the other. 
We present state-of-the-art results for depth super-resolution tasks as well as visually attractive, enhanced generated 3D models.
Our main focus is on situations in which there is no large-scale accurate ground-truth depth map to train on, since the depth maps are captured by a low-cost scanner, making it either noisy or quantized.
\end{abstract}
% IEEEtran.cls defaults to using nonbold math in the Abstract.
% This preserves the distinction between vectors and scalars. However,
% if the journal you are submitting to favors bold math in the abstract,
% then you can use LaTeX's standard command \boldmath at the very start
% of the abstract to achieve this. Many IEEE journals frown on math
% in the abstract anyway.

% \twocolumn[{%
% \renewcommand\twocolumn[1][]{#1}%
% \maketitle
% \begin{center}
%     \centering
%     \includegraphics[width=1.0\textwidth]{images/Top_figure_new_bed_cropped.png}
%     \captionof{figure}{Refined depth for an image taken from the Matterport3D data-set \cite{Matterport3D} displayed as a mesh in 3D space. From left to right: RGB image, depth input at 8-bit (simulating a low-cost scanner), our geometrically enhanced depth, ground-truth at 16-bit. We were able to retrieve missing geometric features not visible to the low-cost depth scanner yet captured in a regular camera. Furthermore, we were able to reconstruct details that are hidden by the noise of the 16-bit ground truth such as the texture of the carpet}
%     \label{fig: TopFig}
% \end{center}%
% }]

\begin{figure*}
    \begin{center}
        \includegraphics[width=1.0\textwidth]{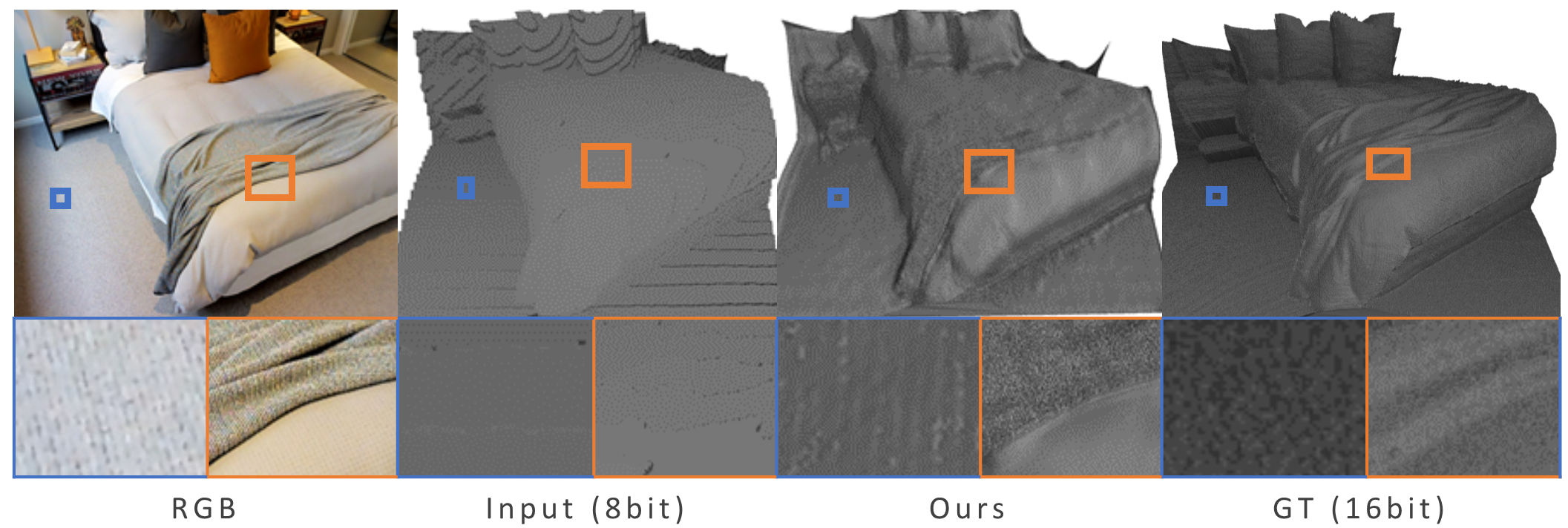}
    \end{center}
  \caption{Refined depth for an image taken from the Matterport3D data-set \cite{Matterport3D} displayed as a mesh in 3D space. From left to right: RGB image, depth input at 8-bit (simulating a low-cost scanner), our geometrically enhanced depth, ground-truth at 16-bit. We were able to retrieve missing geometric features not visible to the low-cost depth scanner yet captured in a regular camera. Furthermore, we were able to reconstruct details that are hidden by the noise of the 16-bit ground truth such as the texture of the carpet}
\label{fig: TopFig}
\end{figure*}

% Note that keywords are not normally used for peerreview papers.
\IEEEpeerreviewmaketitle

\section{Introduction}

RGBD cameras are important in numerous applications spread across multiple fields such as robotics \cite{surveyDepthARVR} \cite{kinectFusion} \cite{AREDepth} \cite{JointDSR}, augmented reality  \cite{AREDepth} \cite{kinectFusion2} \cite{PDSR} \cite{DSR_Stereo}, 
 \cite{surveyDepthARVR},  augmented reality \cite{kinectFusion} \cite{AREDepth} \cite{kinectFusion2} \cite{DSR_Stereo}
and human-computer interaction \cite{kinectFusion} \cite{AREDepth} \cite{kinectFusion2}
. They provide fast and reliable depth maps that allow us to distinguish between objects and background, and to have a better understanding of our environment. Unfortunately, due to physical limitations of the sensors, the resulted depth maps are usually at a lower resolution than the corresponding RGB image \cite{PhotoDSR} \cite{KimmelSpecularSR} \cite{Single_shot_sr}
and are often interpolated in a simplistic manner to provide an output that seems to have a higher resolution. In addition, when using a low-cost sensor the depth pixels contain missing values and are either highly quantized or noisy.
Several approaches are commonly used in order to improve depth maps. 1) super-resolution (SR) of the depth map without additional information \cite{PDSR} \cite{PMBANet} \cite{RGDR} \cite{MSFrequencyDSR}
; These methods are usually divided into two subcategories: the first is filter-based methods \cite{BilateralFilt} \cite{OtherBilateralFilt}
, that are very fast and simple, however, generate blurring and artifacts. The second is optimization-based methods \cite{MRF_DSR} \cite{IntExt}
, which design complex regularization or explore prior to constrain the reconstruction of important high-resolution (HR) depth map. These methods require solving the global energy minimization problem which is time-consuming \cite{HierarchicalDSR} \cite{EdgeDSR}
2) depth prediction from a single RGB image; These methods face a different problem since there are infinite distinct 3D scenes that can produce the same 2D RGB image \cite{OrdinalMonoDepth} \cite{DiggingMonoDepth} \cite{bts}
, meaning that reconstructing the 3D scene from the RGB image requires finding the inverse of a non-injective function.
 3) RGB information fused with the quantized low-resolution (LR) depth in order to produce a finer, more detailed depth map. The RGB image is usually available and containing HR information \cite{PDSR} \cite{PMBANet} \cite{RGDR} \cite{MSFrequencyDSR}.

 In this work, we tackle depth enhancement task using new architecture for image-depth fusion. We report state-of-the-art results on known benchmarks but even more appealing is our ability to transfer fine details from images to depth, generating 3D models that better explain the geometry going beyond the physical limitation of the depth sensor. To use known benchmarks that have been captured using expensive scanners and simulate a low-cost sensor, we quantized the depth map into 8-bits and used it as our only depth map.

\subsection{Contributions}

Our contributions are three-fold:
\begin{itemize}
  \item Introduce the first self-supervised architecture for geometric enhancements guided by visual content.
  \item Propose new cost functions for soft data-fusion.
  \item We present a zero-shot depth super-resolution method based on a DNN that requires no prior training and can be optimized from a single RGBD image. 
\end{itemize}

\section{Related Work}

We identify three different research lines of work for depth multi-resolution: Super-resolution from depth; Image to depth, usually from a single image; and Color guided depth super-resolution. We focus on the latter.

%-------------------------------------------------------------------------
\subsection{Depth super resolution} 
Under this umbrella, we distinguish between axiomatic methods and learning-based~\cite{HierarchicalDSR} \cite{EdgeDSR}, usually deep learning techniques \cite{PDSR}.
Image super-resolution \cite{BlindSuperResolution}, which is an extremely popular technique, inspired many depth SR algorithms, where the network remains almost identical but the trained data is replaced with relevant sampling. \cite{SRandDenois} created a method that specifically fits depth maps and used for simultaneous SR and denoising. They use a coupled dictionary learning method with locality coordinate constraints to reconstruct the high-resolution depth map. \cite{mandal2016depth} proposed a method that combines SR with point-cloud completion for upsampling of both uniform and nonuniform grids. \cite{ferstl2015variational} combines recent learning-based up-sampling with variational SR. First, they trained a high-resolution edge prior to using an external data-set, and then they used this as guidance in the pipeline.
Due to the lack of additional information that exists in the RGB image, none of these methods can reconstruct small structures that were lost during the acquisition process. The distinction between real-world edges and edges that results from artifacts is also more difficult and may cause errors.

%-------------------------------------------------------------------------
\subsection{Image to depth estimation}
The physical limitations of depth sensors raised an important question - can we predict depth from a single image given enough data just from viewing an image. Surprisingly, when the networks are rich enough and trained on enough data the answer is yes in many scenarios \cite{DIML}. \cite{RankingLossDepthPrediction} created a pair-wise ranking loss and a sampling strategy. Instead of randomly sampling point pairs, they guide the sampling to better characterize structure of important regions based on the low-level edge maps and high-level object instance masks. \cite{bts} uses densely encoded features, given dilated convolutions, in order to achieve more effective guidance. To that end, they utilize novel local planar guidance layers located at multiple stages in the decoding phase.
\cite{OrdinalMonoDepth} uses dilated convolutions and avoids down-sampling via spatial pooling. This allows them to obtain HR feature information. In addition, they trained their network using an ordinary regression loss, which achieved higher accuracy and faster convergence.
Unlike these methods, we base our prediction on a lower resolution depth map that provides us with a decent estimation of depth. This solves one of the greatest issues of depth estimation - perspective.
%-------------------------------------------------------------------------
\subsection{Color guided depth super-resolution}
Images provide important details for inferring the geometry. The direction of the normal, for example, is correlated to color changes in the image \cite{PDSR}. 
The challenge here is that the modality is different, and similar depth might be represented in an image with different color values, and on the other hand, different depth might have similar color.  
Among the popular solutions we find, \cite{RGDR} which created an optimization framework for color guided depth map restoration, by adopting a robust penalty function to model the smoothness term of their model.
\cite{PMBANet} formulate a separate guidance branch as prior knowledge to recover the depth details in multiple scales.
\cite{Single_shot_sr} combine heterogeneous depth and color data to
tackle the depth super-resolution and shape-from-shading problems. \cite{li2020depth} created a correlation-controlled color guidance block, using this block, they designed a network that consists of two multi-scale sub-networks that provide guidance and estimate depth. 
Last, in order to create photorealistic depth maps that fit the real world, \cite{PDSR} created a loss function that measures the quality of depth map upsampling using renderings of resulting 3D surfaces.

\subsection{Deep zero-shot models}
Lack of training data raised new interesting problems referred to as few-shot-learning and zero-shot-learning. in layman's terms not only we do not have labels, but we have not seen any sampled data from the distribution examined.
One popular way to achieve that is to allow the network to train directly in test time, optimizing for a specific end case.
 \cite{SinGAN} created an unconditional image generator that can be used for a variety of tasks, such as paint-to-image, editing of images, and super-resolution. \cite{zero-shot-SR} is used for super-resolution of natural images by exploiting the internal recurrence of information inside a single image. \cite{TextureGAN} created a non-stationary texture synthesis method based on a generative adversarial network (GAN). \cite{meta_transfer_zero_shot} exploit both external and internal information for zeros-shot super-resolution, where one single gradient update can yield considerable results. In our work, we show that our loss functions are robust enough to train on the provided image during test - and provide highly appealing results.

\section{Proposed Method}

Our method focuses on super-resolution and reconstruction of fine details that were lost due to the imperfect acquisition process of the depth map. Simultaneously, we fix the depth artifacts created on smooth areas, we refer to as stairs artifact which is a byproduct of sensors' binning, displayed in figures \ref{fig: TopFig} and \ref{fig:FE}.
To that end, we designed a depth super-resolution (DSR) module that takes as input an $H \times W$ depth map with the corresponding $2H \times 2W$ gray-scale image and output a refined $2H \times 2W$ depth map, shown in figure \ref{fig:DRB}. We use this module sequentially three times, to get an $\times8$ resolution gain, shown in figure \ref{fig:FullNet}. Hence, the network can be formulated as:
\begin{equation}
D_{HR} = f(I_{HR}, f(I_{DS2}, f(I_{DS4}, D_{LR})))
\end{equation}

Where $D_{HR}$ is the predicted HR depth, $D_{LR}$ is the low-resolution depth, and $I_{DSx}$ is the RGB image down-sampled $x$ times.

\subsection{Architecture}

\begin{figure*}
    \begin{center}
        \fbox{\includegraphics[width=0.95\textwidth]{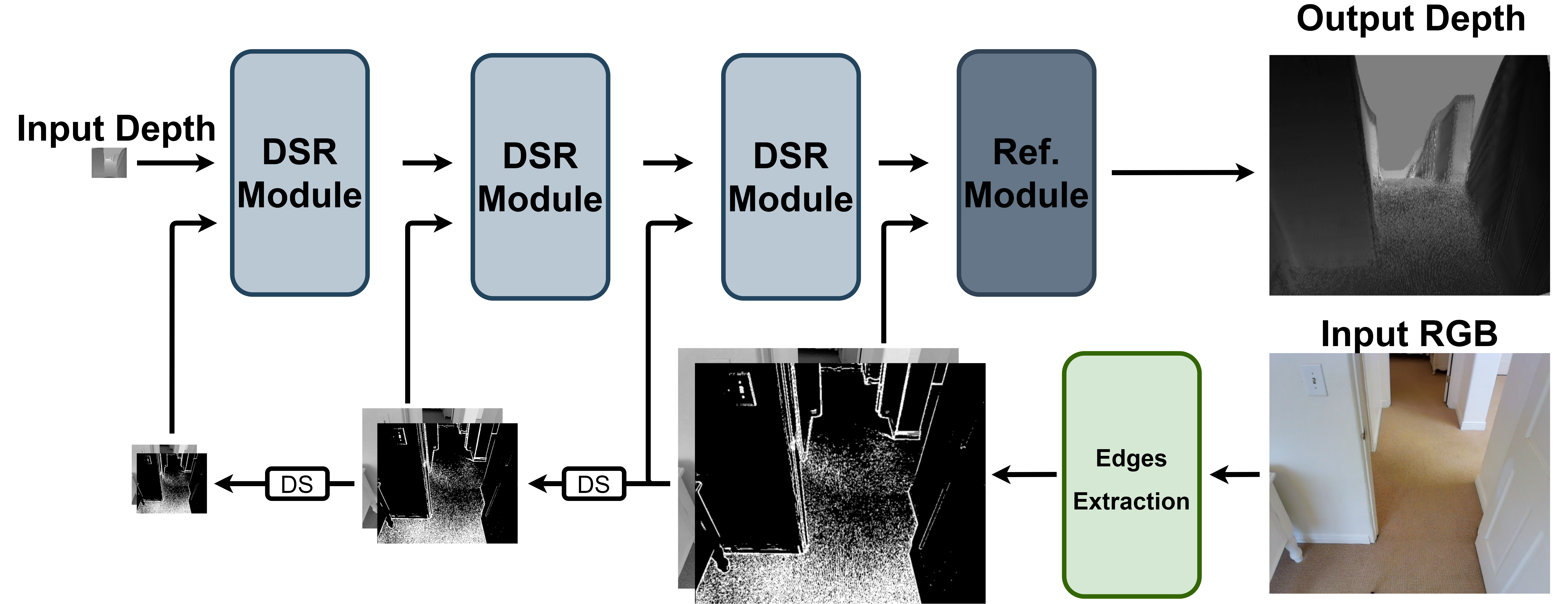}
        }
\end{center}
  \caption{Full Network Architecture: The network takes as input the HR RGB image, at a resolution on $2^{n}H$ by $2^{n}W$ and LR depth map at a resolution of $H$ by $W$. After the RGB image passes the Edge extraction module, returning the edge map concatenated to the gray-scale image, it is down-sampled (DS) $n-1$ times, each time at a factor of 2. Each down-sample is the input of a different DSR module as well as the corresponding depth map.}
\label{fig:FullNet}
\end{figure*}

Our architecture uses three consecutive upsampling modules called Depth Super-Resolution, DSR in short, each one up-sample the resolution by a factor of $\times2$ and refines the depth. Then, a refinement module, identical to the one inside each DSR module, is used for fine-tuning before returning the final output as shown in figure \ref{fig:FullNet}.
Our modular design allows up-sampling by different factors - stacking $n$ DSR modules resulting in $2^{n}$ resolution increase.
%For different resolution increases, the DSR modules can be used n times and return $2^{n}$ times resolution increase.

\subsubsection{DSR module}
\begin{figure}
\centering
    \includegraphics[width=0.75\textwidth]{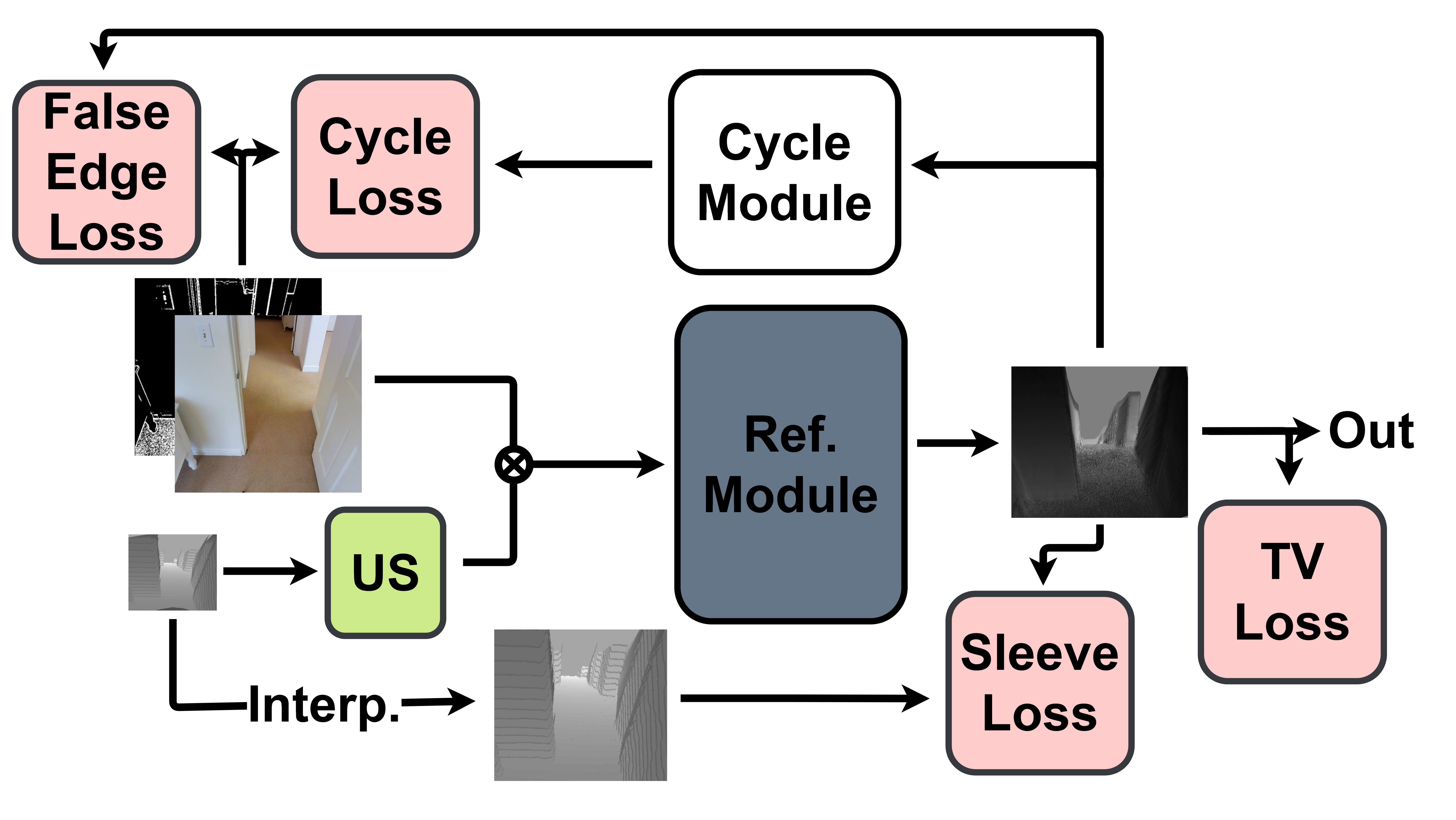}
    \caption{Depth Super-Resolution (DSR) Module: the lower resolution depth passes through an up-sample (US) module that generates high resolution feature map. The feature map is concatenated to the edges and passes the refinement module to create the high resolution depth output.}
\label{fig:DRB}
\end{figure}

Each DSR module, displayed in figure \ref{fig:DRB}, receives two inputs. The first is the higher ($\times2$) resolution grayscale and image with concatenated edges for enhancing fine details, and the second is the LR depth map. The depth input passes through two $3\times3$ convolutional layers, a $3\times3$ deconvolution layer, that upsamples the depth to the height and width of the RGB image, and another three $3\times3$ convolutional layers. At this point, when both inputs are at the same resolution, we concatenate the two channels.

The last unit in the pipeline is a refinement module. It starts with a Dilated Inception Module, inspired by ~\cite{DiletedInception} and~\cite{diletedInceptionSR} with dilation rates of [1, 2, 4, 8]. Then, ten convolutional layers, where the last one is a $1\times1$ convolution, that return a single channel depth map, used as input for the next block and for loss calculation for this block.

\subsubsection{Cycle module} 
\label{cyc_mod}
One of the main challenges we face here is the lack of reliable labels. As the title implies, we do not really have ground-truth. To overcome this limitation, we suggest transferring information from the geometry space back to the image space. But, unlike other transformer networks, we only care about the high frequencies or edges in this case. Specifically, this module
starts by applying edge extraction to the predicted depth map. The edges are concatenated to the predicted depth and together, they pass our geometry-to-image transformer. This module, applied for each DSR output, is constructed from 15 consecutive $3\times3$ convolutions and ends with a single $1\times1$ convolution layer.

\subsubsection{Edge extraction}
\label{edg_mod}
The edges extracted from the color image contain important information that is required for quality refinement and SR of depth maps. However, since the RGB image and the depth map are two different modalities, the magnitude of the derivatives can be washed out.
To face this issue we start by transforming the RGB image into a grayscale image. Then, we used the Sobel operator for edge extraction. Next, by thresholding the magnitude of the operator, using the $p$ percentile ($p$ is set to $50$ by default) in that image as a threshold, we created a binary map that shows "edge" vs "no edges" according to the RGB image.

%-------------------------------------------------------------------------
\subsection{Losses}
The loss functions used during training were crafted especially to fit the task of self-supervised depth refinement, under the premise that although the ground truth is absent, it can be approximated with a certain error factor by a simple interpolation of the input. Our network calculates loss for each scale and penalizes known artifacts that exist in quantized depth maps, such as "Stairs Artifact". Specifically, we introduce four different losses. Sleeve loss \ref{SL}, which provides some level of tolerance, not punishing at all if the solution close enough to the interpolated quantized depth. Cycle loss \ref{CL}, making sure fine details appear in the geometry. False edge loss \ref{FE}, which removes geometric artifacts that appear in the interpolated quantized depth. Last, Total Variation loss \ref{TV}, for a smooth solution. 
Let us elaborate on each loss function.

\subsubsection{Sleeve loss} \label{SL}
After approximating the ground truth depth data with the up-sampled lower resolution depth map, we train the network in a manner that allows changes and corrections of the input channel even in locations we received data from the depth sensor.
Here we propose to use a sleeve loss, defined by 

\begin{equation}
  L_{sleeve} = max(|Y - \hat{Y}|-s, 0)
\end{equation}

Where $Y$ is an interpolated depth from the sensor, $\hat{Y}$ is the predicted depth and $s$ is the allowed error sleeve.

This formulation allows the network to detach its output from the approximated ground truth without any penalty up to a given threshold - $s$. Due to the native smoothness effect of convolutional neural networks not only do we get a smooth solution, but it falls within a pipe shape along with the scanned, binned, depth data.
 The obvious drawback of using this loss function is over-smoothness, and indeed fine details are not recovered by this one. Here we set the low frequencies of the domain on one hand, and force it to correlate with the depth data we received as an input on the other hand.

%-------------------------------------------------------------------------
\subsubsection{Cycle loss}
\label{CL}
Due to the limitations of current low-cost scanners, and the difficulty in acquiring large-scale data-sets for supervised training we chose to focus on unsupervised solutions - learnable and zero/few-shot. To achieve this goal, motivated by the cyclic loss presented in ~\cite{Zhu_2017_ICCV}, we have built a transformer network \ref{cyc_mod}, that transfers the edges of the predicted depth into the domain of the edges extracted from a grayscale image. These edges are compared to the genuine edges of the grayscale image, extracted using the above-mentioned Edge extraction module \ref{edg_mod}. 
We chose to cycle back and predict only the edges and not the image since it is a simpler task, and focus just on the necessity required to guide the depth reconstruction unit.
The comparison is done as a simple $L_1$ norm.
%-------------------------------------------------------------------------
\subsubsection{False-Edge loss}
\label{FE}
Due to our choice of approximating the ground truth by interpolating the LR quantized depth map, our approximation contains many false edges. Any smooth surface at an inclined slope contains the above-mentioned Stairs Artifact. To deal with this issue, we created the False Edge loss that relies on the premise that edges that are reliable to the scanned scene can be found in both the depth map and the grayscale image. We use our edge extraction module to find the edges in the grayscale image. Then, we sum the edges of the predicted depth that correspond to smooth areas on the RGB image, meaning edges that are found in the depth map but not in the grayscale image. Calculated as follows:

\begin{equation}
  L_{fe}(d, gs) = \sum {E_{d}(1-E_{gs})}
\end{equation}

Where $d$ and $gs$ are the depth map and grayscale image, $E_{d}$ and $E_{gs}$ are the edges extracted from the depth map and grayscale image accordingly.

\begin{figure}
\centering
        \includegraphics[width=0.75\textwidth]{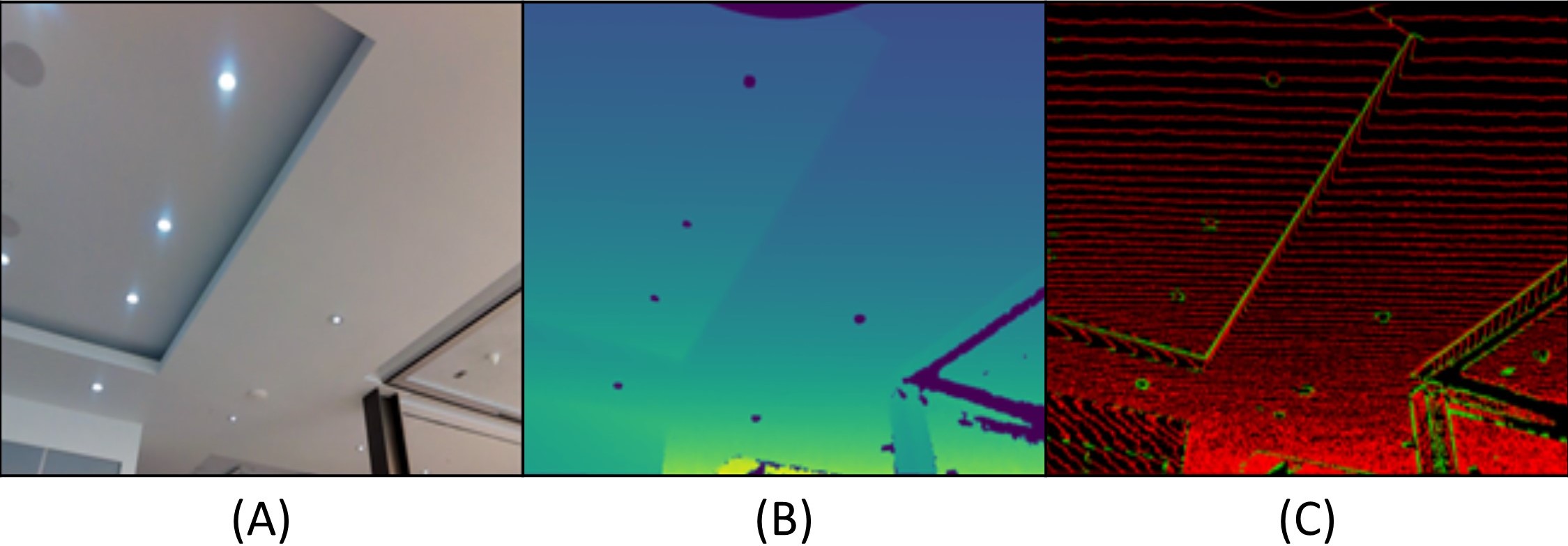}
  \caption{Example of false and true depth edges. A) is an RGB image of a ceiling. B) Is the corresponding depth map, and C) are the quantized depth map edges in which green marks true edges and red are false edges created by the stairs artifact.}
\label{fig:FE}
\end{figure}

Figure \ref{fig:FE} shows a visual representation of this phenomena, the green edges are edges that fit the RGB image (around the lights and the rectangle created by the drop ceiling), and as a result will not yield loss. The red edges are the result of the stairs-artifact and do not appear in the RGB image nor the real world. The summation of these edges is the definition of the False-Edge loss. Differently from the Cycle-loss, which compares edges in the image domain, this loss is applied at the geometry domain and only punishes false edges (edges that exist where edges should not be found and are marked in red. This loss does not punish the absence of true edges (edges that exists where edges should be found and are marked in green).

\subsubsection{Total-Variation regularization}
\label{TV}
In order to force smoother predictions, we added a total variation (TV) regularization. Similarly to \cite{aly2005image}, formulated as:
\begin{equation}
  R_{tv}(d) = \sum_{i, j} (\sqrt{|d_{i, j} - d_{i+1, j}|^{2} + |d_{i, j} - d_{j+1, i}|^{2}} )
\end{equation}
where $R_{tv}(d)$ is the TV regularization value for the depth map $d$ and $d_{i, j}$ is the depth pixel at the $(i, j)$ location.

\begin{figure}
\centering
        \includegraphics[width=0.95\textwidth]{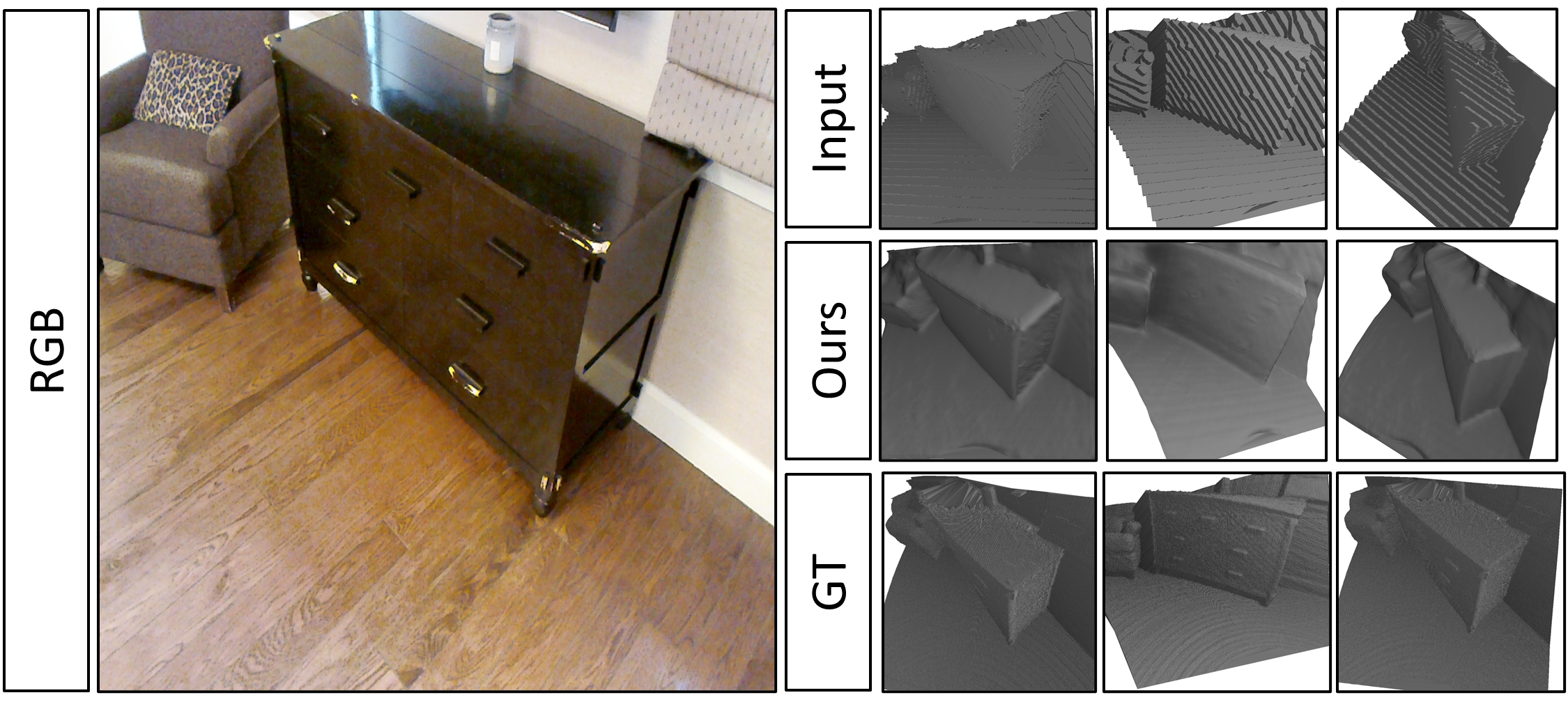}
  \caption{Results for multi-view test: presenting our results in a natural view for AR/VR environment. Not only we remove the quantization artifacts, but we also keep edges and corners of the furniture.}
\label{fig:ResultsMultiVew2}
\end{figure}

\begin{figure*}
    \begin{center}
        \includegraphics[width=0.95\textwidth]{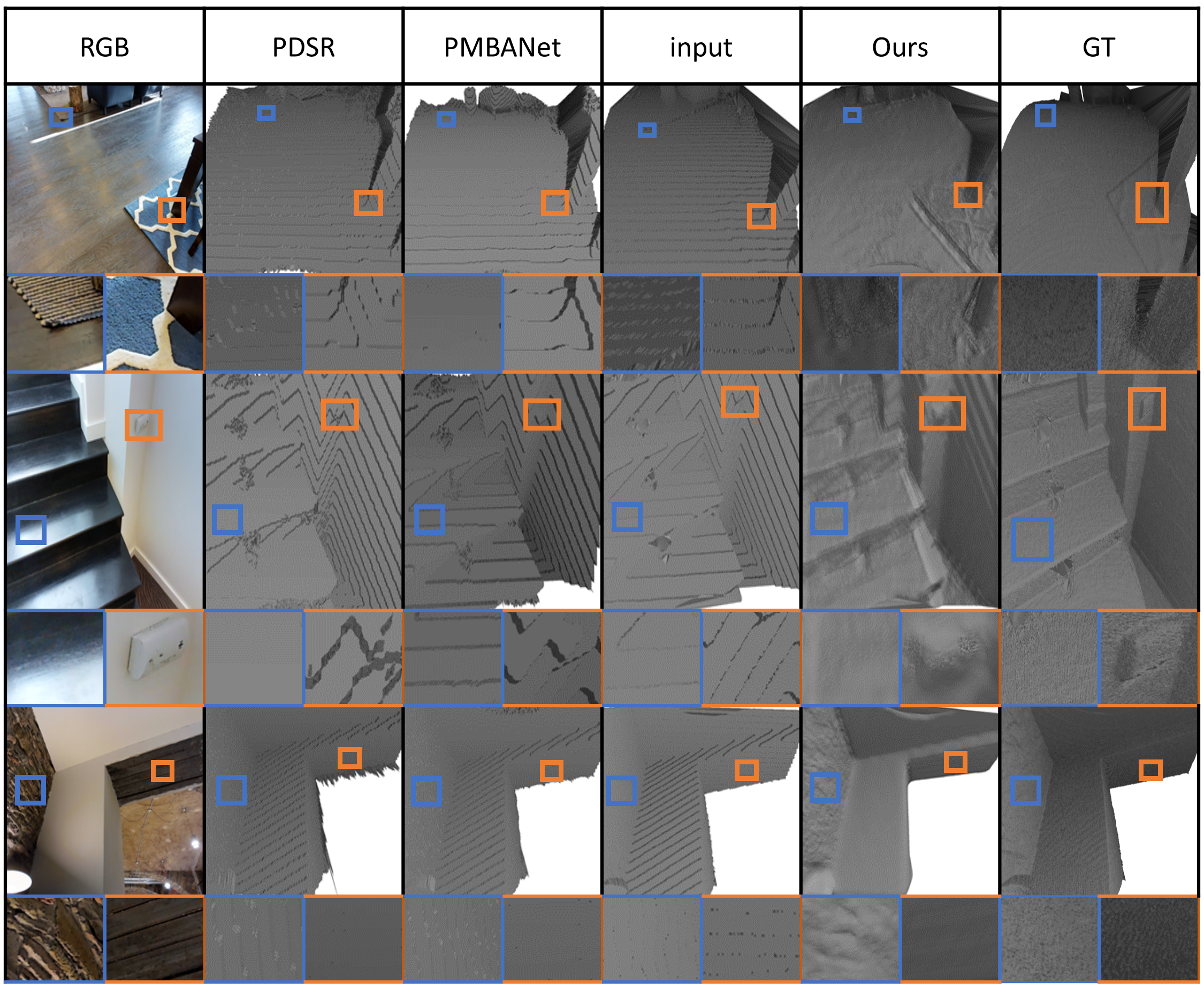}
    \end{center}
  \caption{Qualitative results for Matterport3D \cite{Matterport3D}: The proposed method was able to retrieve curved shapes, edges, and corners while smoothing only relevant parts in the provided depth, restoring even fine details that were lost in the noise of the ground truth such as the carpet on the top row or the texture of the wall and the wood on the bottom row.}
\label{fig:ResultsMP}
\end{figure*}

% \begin{figure*}
%     \begin{center}
%         \includegraphics[width=1.0\textwidth]{images/MP_table_new_cropped.png}
%     \end{center}
%   \caption{Qualitative results for Matterport3D \cite{Matterport3D}: The proposed method was able to retrieve curved shapes, edges, and corners while smoothing only relevant parts in the provided depth, visually surpassing the ground truth.}
% \label{fig:ResultsMP}
% \end{figure*}

\begin{figure*}
    \begin{center}
        \includegraphics[width=0.95\textwidth]{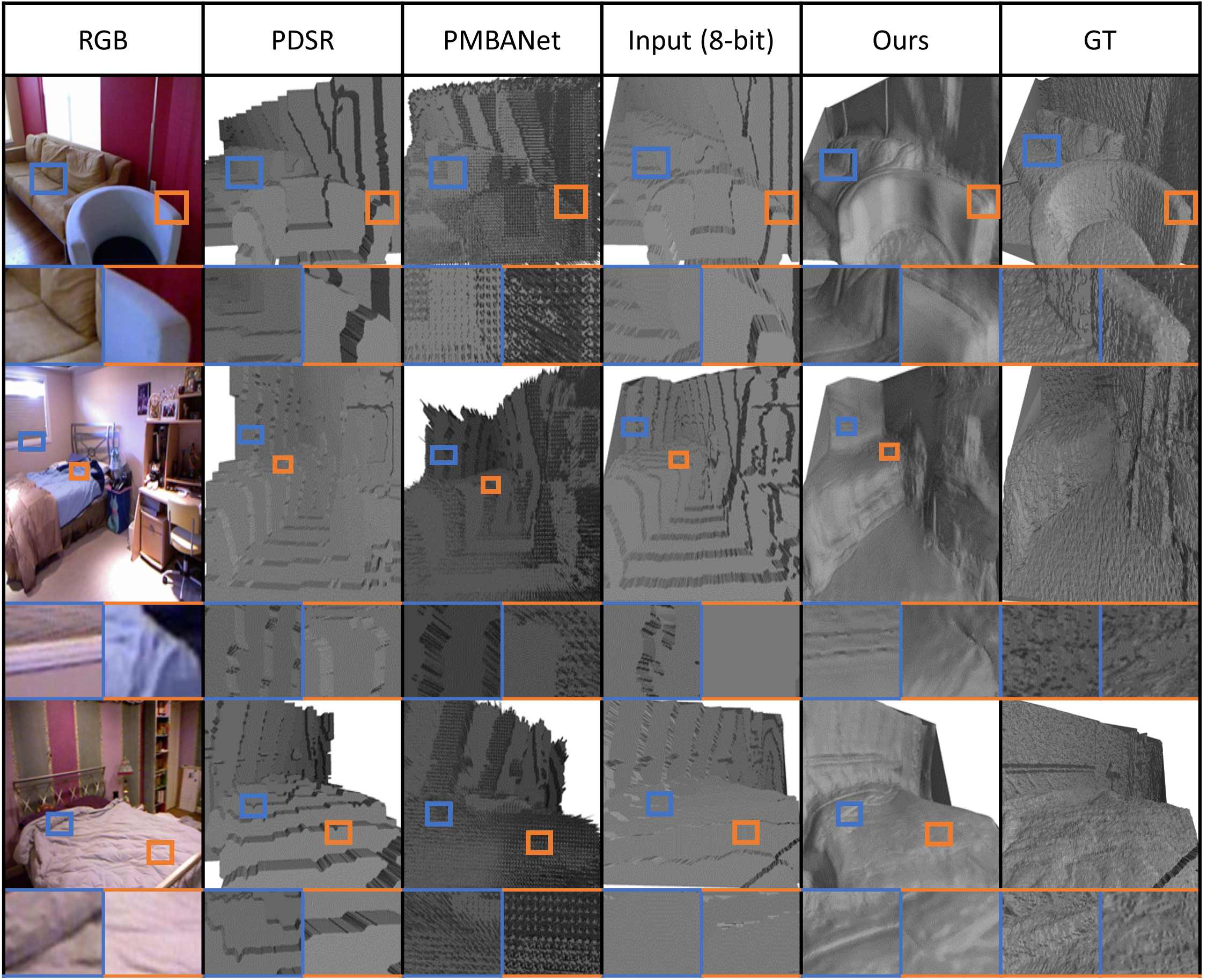}
    \end{center}
  \caption{Qualitative results for NYUv2 \cite{NYUv2}: Reconstructed depth information for in-doors scenes. The proposed method was able to retrieve curved shapes, edges, and corners while removing artifacts and smoothing only relevant parts. It can be seen that although the ground truth contain noise, especially visible on smooth surfaces, the predicted depth is smoother and cleaner.}
\label{fig:ResultsNYUv2}
\end{figure*}

\subsection{Implementation details}
Our architecture is built out of three DSR blocks followed by a refinement module. Each DSR block contains one cycle module and one Refinement module.

% The weights between the different losses were tweaked manually, more details can be found in the supplementary materials.

The zero-shot training is accomplished simply by optimizing the network on one image without any changes aside from the weights between the different losses.

We chose to construct each module in the following way: All layers (both convolutional and deconvolutional) have 8 output channels, with exception of the final two layers in each block/module. The second to last layer outputs $k\geq8$ channels ($k$ is a hyperparameter, set to 16 by default) and the last layer outputs $1$ channel (meant to fit the dimensions of the output to the dimensions of the depth map). Hence, for each given module the number of filters for each layer will be as follows: [8, 8, 8, ..., 8, $k$, 1].

For optimization, we chose Adam optimizer with $\mu = 0.9$, $\beta_{1} = 0.9$, $\beta_{2} = 0.99$ and $\epsilon = 10^{-8}$. The initial learning rate is set to $10^{-4}$ and decreased by 0.5 every 10 epochs. Zero-shot training was done with the same parameters.
We trained the network using a single NVIDIA RTX 2080ti. 

For each visual test, we divided each data-set for 70\% train and 30\% test. We used the HR depth maps as ground truth. For the input of the networks, we downsampled the HR depth maps $\times8$ using Nearest-Neighbor (NN) interpolation, then converted the result to 8 bit. Similarly to other works, we used several common metrics to evaluate our performance. Specifically, we used 
Mean Square Error (MSE), Root Mean Square Error (rMSE), Mean Absolute Error (MAE), Absolute Relative Error (ARE), Peak Signal-to-Noise Ratio (PSNR), and Median Error (ME). 

%-------------------------------------------------------------------------

\setlength{\tabcolsep}{4pt}
\begin{table}
\begin{center}
\caption{Numerical results for ETH3D stereo dataset \cite{ETH3D}: This table presents the MSE, rMSE, MAE, absolute relative error (ARE), PSNR, and median error of our network compared to different methods (by running their published code and using pre-trained weights when those are given). We outperform all methods by a large margin.}
\label{table:TableETH3D}
\begin{tabular}{llllllll}
\hline\noalign{\smallskip}
Method && MSE & rMSE & MAE & ARE & PSNR & Median Err. \\
\noalign{\smallskip}
\hline
\noalign{\smallskip}
earest && 0.904 & 0.819 & 0.763 & 0.299 & 26.953 & 0.504 \\
 
Bilinear && 0.895 & 0.814 & 0.762 & 0.298 & 27.013 & 0.503 \\
 
RGDR~\cite{RGDR} && 0.767 & 0.78 & 0.728 & 0.277 & 26.962 & 0.702 \\

PMBANet~\cite{PMBANet} && 0.806 & 0.799 & 0.734 & 0.281 & 26.883 & 0.737 \\

PDSR~\cite{PDSR} && 0.635 & 0.69 & 0.639 & 0.250 & 28.286 & 0.631 \\

 \textbf{Ours} && 0.205 & \textbf{0.278} & 0.173 & \textbf{0.053} & 39.5 & \textbf{0.046} \\ 
 
 \textbf{Ours} (zero-shot) && \textbf{0.196} & 0.280 & \textbf{0.172} & 0.054 & \textbf{39.903} & 0.05 \\

\hline
\end{tabular}
\end{center}
\end{table}
\setlength{\tabcolsep}{1.4pt}

%%%%%%%%%%%%%%%%%%%%%%%%%%%%%%%%%%%%%%%%%%%%%%%%%%%%%%%%%%%%%%%%%%%%%%%%%%%%%%%%%%%%%%%%%%%%%%%%

% \begin{center}
% \begin{table*}
%  \begin{tabular}{p{0.2\linewidth}|p{0.005\linewidth}p{0.115\linewidth}p{0.115\linewidth}p{0.115\linewidth}p{0.115\linewidth}p{0.115\linewidth}p{0.115\linewidth}}

% \toprule
% Method && MSE & rMSE & MAE & ARE & PSNR & Median Err. \\ [1.0ex] 

% \toprule
%  Nearest && 0.904 & 0.819 & 0.763 & 0.299 & 26.953 & 0.504 \\ [1ex]
 
%  Bilinear && 0.895 & 0.814 & 0.762 & 0.298 & 27.013 & 0.503 \\ [1ex]
 
% \midrule
%   RGDR~\cite{RGDR} && 0.767 & 0.78 & 0.728 & 0.277 & 26.962 & 0.702 \\ [1ex]

% PMBANet~\cite{PMBANet} && 0.806 & 0.799 & 0.734 & 0.281 & 26.883 & 0.737 \\ [1ex]

% PDSR~\cite{PDSR} && 0.635 & 0.69 & 0.639 & 0.250 & 28.286 & 0.631 \\ [1ex]

% \midrule
%  \textbf{Ours} && 0.205 & \textbf{0.278} & 0.173 & \textbf{0.053} & 39.5 & \textbf{0.046} \\ [1ex]
 
%  \textbf{Ours} (zero-shot) && \textbf{0.196} & 0.280 & \textbf{0.172} & 0.054 & \textbf{39.903} & 0.05 \\ [1ex] 
%  \bottomrule

% \end{tabular}
%   \caption{Numerical results for ETH3D stereo dataset \cite{ETH3D}: This table presents the MSE, rMSE, MAE, absolute relative error (ARE), PSNR, and median error of our network compared to different methods (by running their published code and using pre-trained weights when those are given). We outperform all methods by a large margin.}
% \label{fig:TableETH3D}
% \end{table*}
% \end{center}

\section{Results}
We evaluated our method quantitatively and qualitatively by refining and enhancing the depth resolution of several common RGBD indoor data-sets. 

\subsection{Data-sets}
\begin{itemize}
\item\textbf{Matterport3D~\cite{Matterport3D}:}
This large and diverse indoors RGB-D data-set was created in 2017 using the Matterport camera. Both the RGB images and depth maps in this data-set are at a resolution of 1280x1024.\\

\item\textbf{NYUv2~\cite{NYUv2}:}
This RGB-D data-set was created in 2012 using a Kinect depth sensor, originally for semantic segmentation tasks. Both the RGB images and depth maps in this data-set are at a resolution of 640x480. The Train/Test split was done using the provided official split.\\

\item\textbf{ETH3D \cite{ETH3D}:} The creation of this data-set was motivated by the limitations of existing multi-view stereo benchmarks. Since many of these limitations are similar to the limitations in many RGB-D data-sets it fits our needs almost perfectly. Both the RGB images and depth maps in this data-set are at a resolution of 6048x4032, the depth is sparse but not quantized. This data-sets depth is very clean, containing nose levels that are barley visible.\\
\end{itemize}

Our tests focused on two different scenarios. The first, qualitative, comparing our results to different methods on datasets for which ground-truth exists, but is either quantized or noisy and does not represent the real world accurately enough. This tests the visual appeal of our results. For that test we chose Matterport3D \cite{Matterport3D} and NYUv2 \cite{NYUv2}. The second, quantitive, we compared our results to other methods numerically using ETH3D \cite{ETH3D} and simulated a scenario in which we cannot capture a reliable dataset and need to either train self-supervised methods or train on different datasets and count on a small enough domain gap.
The former test whether the model represents the human expectation of the real-world depth, while the latter, which requires nearly perfect depth maps, allows us to measure whether the model represents the real-world depth accurately.

% this scenario involves testing on a data-set with quantized input and no available ground truth for training.

\subsection{Qualitative results}
 For this evaluation, we used Matterport3D \cite{Matterport3D} and NYUv2 \cite{NYUv2}. These data sets allow us to demonstrate our results in a variety of scenes, taken from widely known sources.
As shown in figures \ref{fig:ResultsMP} and \ref{fig:ResultsNYUv2} Our results are more photo-realistic than the results of other methods and often better than the ground-truth. We restore many of the fine details that were lost during quantization and did not exist in networks' input depth. Additionally, it is visible that the supervised methods recreated the artifacts that existed in the input, and by doing that, generated a less realistic depth map that represents the real world unreliably.
For the second qualitative evaluation, we rotated the mesh created by the projected depth map. This allows us to demonstrate how the results can reliably represent the real world, and that details are restored in a reliable manner that fits the human expectation of depth, see figure \ref{fig:ResultsMultiVew2}. A comparison between zero-shot and unsupervised training can be seen in figure \ref{fig: TopFig} and the supplements.
It is clear that while the main focus of some recent works is to create sharp outputs with undisturbed edges, smooth depth transitions across planes are not a major factor, since this does not occur in the ground truth.

\subsection{Quantitive results}
These tests were done on indoor scans from the ETH3D data-set \cite{ETH3D}. This data set's sparse, yet clean and unquantized depth allowed us to create a low-resolution quantized depth input and test the model's performance using reliable ground truth. The depth maps in this data set contain fine details and minimal levels of noise, making it as close as possible to the scanned scene. This allows us to compare our method's results to other top-rated super-resolution methods, under a scaling factor of $\times8$, as shown in table \ref{table:TableETH3D}. To compare our results to supervised methods under the scenario in which no reliable ground truth exists, we used pre-trained models that were trained on other data sets. In this tested scenario, our model reaches superior results compared to all others. We also show comparable results between our trained and zero-shot methods.

\subsection{Ablation Study}
In order to show the importance of each element of the network, we conducted an ablation study. When possible, we removed the tested element completely, for example, the cycle loss. The sleeve loss was not completely removed, since its absence did not allow the network to converge to any sort of displayable results, so its ablation study was by replacing it with $L1$ (absolute difference) loss. 
The results in table \ref{table:Ablation} show superior results for the complete model.

% \begin{center}
% \begin{table}
% \centering
%  \begin{tabular}{p{0.35\linewidth}|p{0.1\linewidth}p{0.2\linewidth}p{0.2\linewidth}}
% \toprule
%  Removed element   && rMSE           & MAE \\ [1.0ex] 
% \toprule
%  Sleeve loss       && 0.376          & 0.250 \\ [1ex]

%  False Edge loss   && 0.293          & 0.228 \\ [1ex]

%  Total Variation   && 0.353          & 0.246 \\ [1ex]
 
%  Cycle loss        && 1.797          & 1.556 \\ [1ex]
%  \midrule
%  Full Architecture && \textbf{0.280} & \textbf{0.172} \\ [1ex]
%  \bottomrule

% \end{tabular}
%   \caption{Numerical results for ablation study.}
% \label{fig:Ablation}
% \end{table}
% \end{center}

% \setlength{\tabcolsep}{4pt}
% \begin{table}
% \begin{center}
% \caption{Font sizes of headings. Table captions should always be
% positioned {\it above} the tables. The final sentence of a table
% caption should end without a full stop}
% \label{table:headings}
% \begin{tabular}{lll}
% \hline\noalign{\smallskip}
% Heading level & Example & Font size and style\\
% \noalign{\smallskip}
% \hline
% \noalign{\smallskip}
% Title (centered)  & {\Large \bf Lecture Notes \dots} & 14 point, bold\\
% 1st-level heading & {\large \bf 1 Introduction} & 12 point, bold\\
% 2nd-level heading & {\bf 2.1 Printing Area} & 10 point, bold\\
% 3rd-level heading & {\bf Headings.} Text follows \dots & 10 point, bold
% \\
% 4th-level heading & {\it Remark.} Text follows \dots & 10 point,
% italic\\
% \hline
% \end{tabular}
% \end{center}
% \end{table}
% \setlength{\tabcolsep}{1.4pt}

\setlength{\tabcolsep}{4pt}
\begin{table}
\begin{center}
\caption{Numerical results for ablation study. Each line represents the results of the network with a different part missing}
\label{table:Ablation}
\begin{tabular}{lllllll}
\hline\noalign{\smallskip}
& MSE & rMSE & MAE & ABS REL & PSNR & Median Err.\\
\noalign{\smallskip}
\hline
\noalign{\smallskip}
 Sleeve loss & 0.364 & 0.376 & 0.250 & 0.078 & 35.80 & 0.254 \\ 

 False Edge loss & \textbf{0.193} & 0.293 & 0.228 & 0.075 & 37.985 & 0.239 \\

 Total Variation & 0.379 & 0.353 & 0.246 & 0.065 & 38.985 & 0.065 \\
 
 Cycle loss & 6.248 & 1.797 & 1.556 & 0.572 & 4.351 & 4.645 \\

 Full Architecture & 0.196 & \textbf{0.280} & \textbf{0.172} & \textbf{0.054} & \textbf{39.903} & \textbf{0.05} \\ 

\hline
\end{tabular}
\end{center}
\end{table}
\setlength{\tabcolsep}{1.4pt}

% \begin{center}
% \centering
% \begin{table*}
%  \begin{tabular}{p{0.15\linewidth}|p{0.115\linewidth}p{0.115\linewidth}p{0.115\linewidth}p{0.115\linewidth}p{0.115\linewidth}p{0.115\linewidth}}

%  \toprule
%  & MSE & rMSE & MAE & ABS REL & PSNR & Median Err. \\ [1.0ex] 
% \toprule
%  Sleeve loss & 0.364 & 0.376 & 0.250 & 0.078 & 35.80 & 0.254 \\ [1ex]

%  False Edge loss & \textbf{0.193} & 0.293 & 0.228 & 0.075 & 37.985 & 0.239 \\ [1ex]

%  Total Variation & 0.379 & 0.353 & 0.246 & 0.065 & 38.985 & 0.065 \\ [1ex]
 
%  Cycle loss & 6.248 & 1.797 & 1.556 & 0.572 & 4.351 & 4.645 \\ [1ex]
% \midrule
%  Full Architecture & 0.196 & \textbf{0.280} & \textbf{0.172} & \textbf{0.054} & \textbf{39.903} & \textbf{0.05} \\ [1ex]
%  \bottomrule

% \end{tabular}
%   \caption{Numerical results for ablation study. Each line represents the results of the network with a different part missing.}
% \label{fig:Ablation2}
% \end{table*}
% \end{center}

\subsection*{Failed cases}
\begin{figure}[!htb]
\centering
    \includegraphics[width=0.75\textwidth]{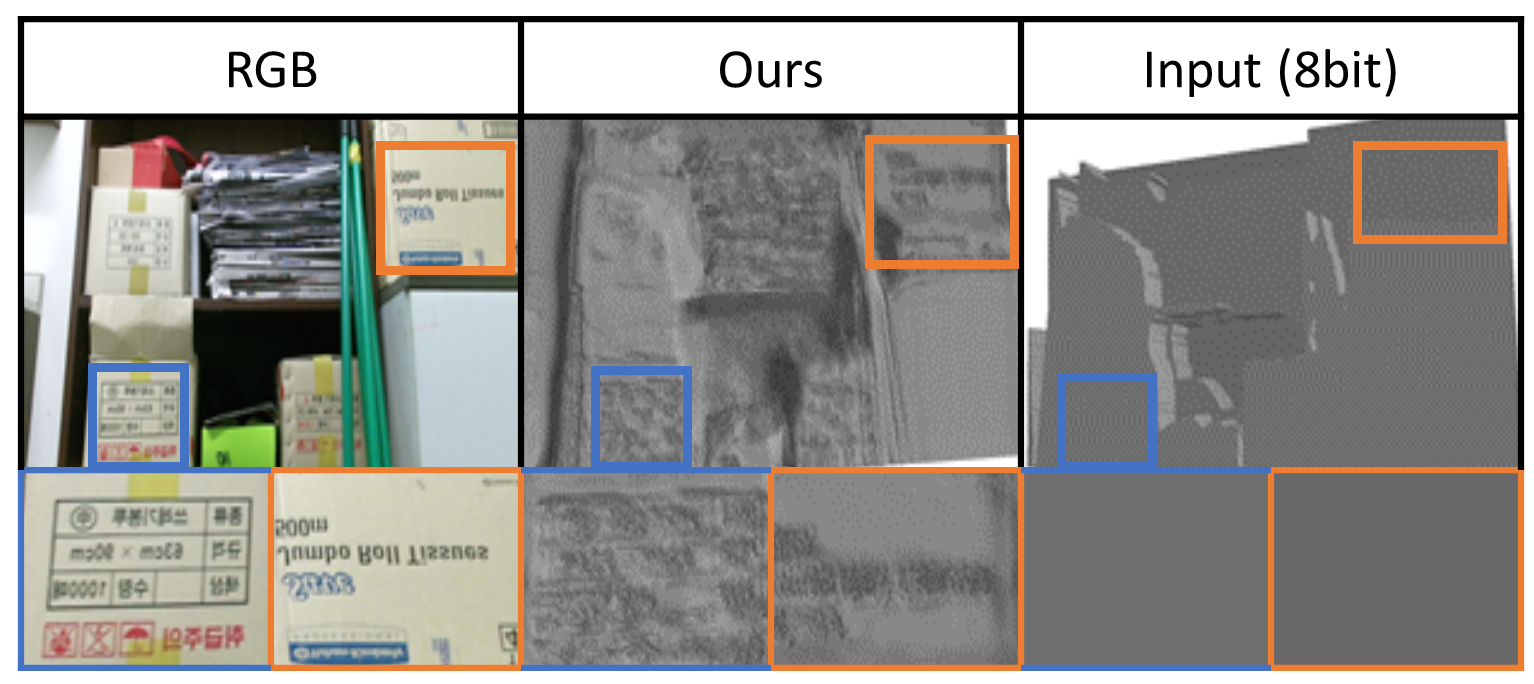}
    \caption{Failure example: Image taken from \cite{DIML}, the text appears in the predicted depth as causing depth changes. This is caused by the strong gradients in the RGB image that the letters create.}
\label{fig:Fail}
\end{figure}

Our network often fails over paintings and writings that cause string edges in the RGB images and does not effect depth.
For dealing with these failed cases we can research the following approach: First, training a different network that segments painting, text, and other strong misleading gradients. Then, we can block them in the RGB image, preventing the gradients from reaching the network.

\section{Conclusion}
We presented the first unsupervised zero-shot neural network architecture for geometric depth refinement and super-resolution. Our method is based on a novel network architecture and loss functions, that not only outperforms known methods on given benchmarks but can retrieve fine details of the geometry that can not be captured in low-cost depth scanners. Due to the lack of other self-supervised depth super-resolution methods, we compared our results to pre-trained supervised methods, trained on different data sets. Under this scenario, the other methods deal with a domain gap that our self-supervised method can avoid. This allowed us to achieve superior results in a variety of tests.

{\small
\bibliographystyle{ieee_fullname}
\bibliography{journal.bib}
}

% \input{6.supplementary}

% For peer review papers, you can put extra information on the cover
% page as needed:
% \ifCLASSOPTIONpeerreview
% \begin{center} \bfseries EDICS Category: 3-BBND \end{center}
% \fi
%
% For peerreview papers, this IEEEtran command inserts a page break and
% creates the second title. It will be ignored for other modes.

\ifCLASSOPTIONcaptionsoff
  \newpage
\fi

\end{document}